\documentclass[letterpaper, 10 pt, conference]{ieeeconf}  
\usepackage{color}
\newcommand{\ie}{i.e.,\ }

\newcommand{\Reffig}[1]{Figure~\ref{#1}}
\newcommand{\Refsec}[1]{Section~\ref{#1}}

\newcommand{\Refeq}[1]{Equation~\ref{#1}}
\newcommand{\Reftab}[1]{Table~\ref{#1}}

\usepackage{graphics} 
\usepackage{epsfig} 
\usepackage{algorithm}
\usepackage{algpseudocode}
\usepackage{amsmath}
\usepackage{amssymb}
\usepackage{subfigure}
\usepackage{graphicx}
\usepackage{subfigure}
\usepackage{booktabs}
\usepackage{threeparttable}
\usepackage{multirow}
\usepackage{multicol}
\usepackage{makecell}
\usepackage{dcolumn}
\usepackage[T1]{fontenc}
\usepackage{diagbox}
\usepackage[normalem]{ulem}
\useunder{\uline}{\ul}{}

\IEEEoverridecommandlockouts                              

\title{\LARGE \bf
LIMOT: A Tightly-Coupled System for LiDAR-Inertial Odometry and Multi-Object Tracking
}

\author{Zhongyang Zhu$^{1}$, Junqiao Zhao$^{*,1, 2}$, Kai Huang$^{3}$, Xuebo Tian$^{1}$, Jiaye Lin$^{1}$, Chen Ye$^{1}$
\thanks{This work is supported by the National Key Research and Development Program of China (No. 2021YFB2501104). \emph{(Corresponding Author: Junqiao Zhao.)}}
\thanks{$^{1}$Zhongyang Zhu, Junqiao Zhao, Xuebo Tian, Jiaye Lin, and Chen Ye are with Department of Computer Science and Technology, 
School of Electronics and Information Engineering, Tongji University, Shanghai, China, and the MOE Key Lab of Embedded System and Service Computing, Tongji University, Shanghai, China 
{\tt\footnotesize (e-mail: 2233057@tongji.edu.cn; zhaojunqiao@tongji.edu.cn; 1930773@tongji.edu.cn;2310920@tongji.edu.cn; yechen@tongji.edu.cn).}}
\thanks{$^{2}$Institute of Intelligent Vehicles, Tongji University, Shanghai, China}
\thanks{$^{3}$Kai Huang is with the School of Surveying and Geo-Informatics, Tongji University, Shanghai, China
{\tt\footnotesize (e-mail: 1911202@tongji.edu.cn).}}
}

\begin{document}

\maketitle
\thispagestyle{empty}
\pagestyle{empty}

\begin{abstract}
Simultaneous localization and mapping (SLAM) is critical to the implementation of autonomous driving.
Most LiDAR-inertial SLAM algorithms assume a static environment, leading to unreliable localization in dynamic environments.
Moreover, the accurate tracking of moving objects is of great significance for the control and planning of autonomous vehicles. 
This study proposes LIMOT, a tightly-coupled multi-object tracking and LiDAR-inertial odometry system that is capable of accurately estimating the poses of both ego-vehicle and objects.
We propose a trajectory-based dynamic feature filtering method, which filters out features belonging to moving objects by leveraging tracking results before scan-matching.
Factor graph-based optimization is then conducted to optimize the bias of the IMU and the poses of both the ego-vehicle and surrounding objects in a sliding window.
Experiments conducted on the KITTI tracking dataset and self-collected dataset show that our method achieves better pose and tracking accuracy than our previous work DL-SLOT and other baseline methods.
Our open-source implementation is available at https://github.com/tiev-tongji/LIMOT.
\end{abstract}


\section{Introduction}
Simultaneous localization and mapping (SLAM) is essential for the operation of autonomous vehicles in Global Navigation Satellite System (GNSS)-denied environments.
Most SLAM systems degrade in dynamic environments because they rely heavily on the assumption of a static environment.
At the same time, multi-object tracking in complex dynamic scenes is crucial for the control and planning of autonomous vehicles. 
Therefore, approaches combining SLAM and multi-object tracking have emerged in recent years \cite{yang2019cubeslam, tian2022dl, lin2023lio-segmot}.
The vast majority of methods perform multi-object tracking and SLAM separately, \ie{loosely-coupled}.
This results in that the tracking accuracy highly depends on ego-pose estimation which is, however, not reliable in dynamic environments.

Recently, tightly-coupled multi-object tracking and vision-based SLAM systems have gained extensive attention \cite{yang2019cubeslam,zhang2020vdo,bescos2021dynaslam,gonzalez2022twistslam}. 
These systems construct a unified optimization framework, in which the poses of both the ego-vehicle and moving objects are jointly optimized.
However, their performance is limited by inaccurate 3D object detection and sensitivity to illumination change and rapid motion. 

In this paper, we propose LIMOT, a tightly-coupled multi-object tracking and LiDAR-inertial odometry system capable of accurately estimating the poses of both the ego-vehicle and surrounding objects. 
First, all movable objects are detected as 3D bounding boxes. 
Simultaneously, inertial measurement unit (IMU) pre-integration is utilized to de-skew LiDAR scans and provide an initial guess for scan-matching.
Then, similar to DL-SLOT \cite{tian2022dl}, a combination of trajectory approximation of tracked objects in a sliding window and the continuous shortest path algorithm \cite{ahuja1988network} is employed to perform data association. 
Object states (stationary or dynamic) can then be determined.

DL-SLOT filters all feature points belonging to movable objects and suffers from feature sparsity when many movable objects are actually static.
However, based on the approximated object trajectories and the estimated motion from IMU pre-integration, feature points belonging to moving objects can be precisely filtered out before scan-matching in LIMOT.
Finally, a factor graph optimization framework is conducted to optimize the bias of the IMU and the poses of both the ego-vehicle and objects in a sliding window.
The main contributions of this work can be summarized as follows:
\begin{itemize}
   \item Development of a tightly-coupled multi-object tracking and LiDAR-inertial odometry system, allowing for joint estimation of the poses of both the ego-vehicle and surrounding objects.
   \item Introduction of a method that leverages the approximated object trajectories to identify and exclude feature points belonging to moving objects, while still utilizing feature points on static movable objects to provide constraints for scan-matching. 
   \item Extensive experiments on different datasets demonstrate the advantages of our system compared to other methods. 
\end{itemize}

\begin{figure*}[ht]
   \centering
   \includegraphics[width=15.0cm]{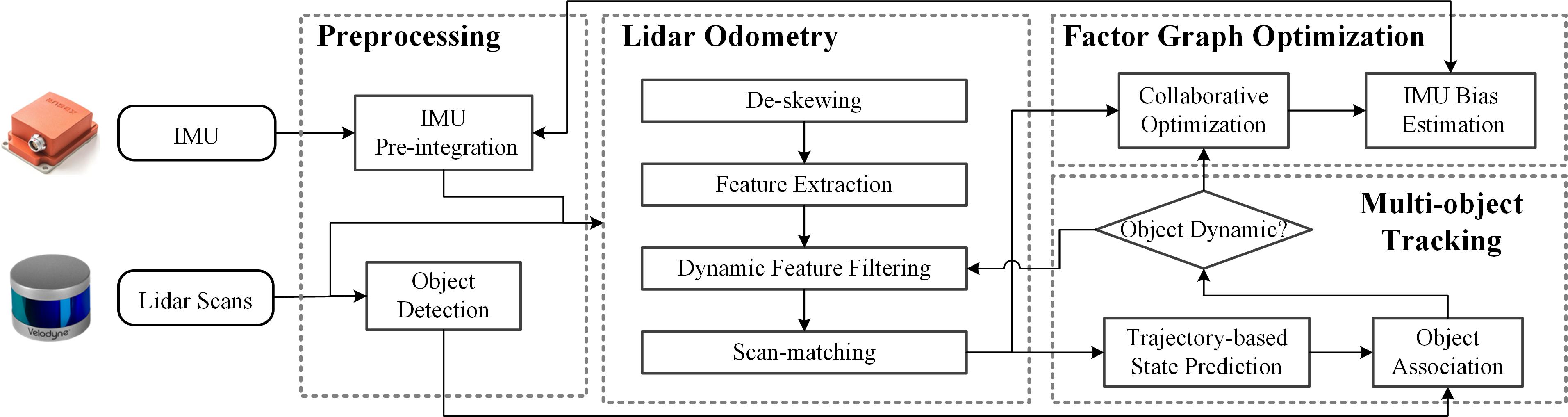}
   \caption{The system architecture of LIMOT. The system consists of the preprocessing, LiDAR odometry, the sliding window-based 3D multi-object tracking, and factor graph optimization.}
   \label{fig_system}
\end{figure*}

\section{Related Works}
In recent years, the multi-object tracking and SLAM system in dynamic environments has been increasingly investigated.
The existing methods can be classified into two types: loosely-coupled and tightly-coupled.

In the former, multi-object tracking and SLAM are performed independently \cite{wang2007simultaneous, ma2022mlo, runz2018maskfusion}.
SLAMMOT\cite{wang2007simultaneous} first proposed simultaneously estimating ego-motion and multi-object motion, establishing a framework to decompose the estimation problem into two separated filter-based estimators. 
\cite{ma2022mlo} estimates ego-motion based on the static background and achieves object tracking through the least-squares method by fusing point clouds and 3D detection.
However, the accuracy of multi-object tracking in these methods is heavily dependent on ego-pose estimation which likely fails in complex dynamic environments.

Tightly-coupled multi-object tracking and SLAM methods are primarily proposed in visual SLAM systems.
Specifically, CubeSLAM \cite{yang2019cubeslam} firstly proposed the use of a multi-view bundle adjustment (BA) to jointly optimize ego-pose, states of objects, and feature points.
Dynamic objects are tracked by a 2D object tracking algorithm \cite{henriques2014high}.
VDO-SLAM \cite{zhang2020vdo} tracks feature points on objects by leveraging dense optical flow. 
Motion model constraints are added to the factor graph to effectively enable the combined optimization of ego-pose and object states.
DynaSLAM II \cite{bescos2021dynaslam} makes use of instance semantic segmentation and ORB feature \cite{rublee2011orb} correspondences to track moving objects, jointly optimizing the static structure and trajectories of the camera and moving objects in a local-temporal window. 
TwistSLAM \cite{gonzalez2022twistslam} achieves data association using optical flow estimation and performs a novel joint optimization by defining inter-cluster constraints modeled by mechanical joints.
These vision-based methods mostly perform tracking in 2D image space and thus suffer from inaccurate 3D object detection, vulnerability to textureless or low-illumination environments, and occlusion. 

Our previous work DL-SLOT \cite{tian2022dl} is, to the best of our knowledge, the first tightly-coupled LiDAR SLAM and multi-object tracking system without IMU.
Recently, LIO-SEGMOT \cite{lin2023lio-segmot}, proposed an optimization framework similar to ours, but without a sliding window. 
As a result, it is computationally expensive when there are multiple objects for tightly-coupled optimization.
In addition, this method does not remove dynamic points, thus exhibiting only slight improvement in pose accuracy compared to LIO-SAM \cite{shan2020lio}. 


\section{Methods}

\subsection{Notation and System Overview}
We consider $W$ as the world frame and ${L_k}$ as the LiDAR frames, related to the $k$-th LiDAR scan at time $t_k$, respectively.
We donate $\mathbf{T}^a_b \in $ SE(3) as the pose of $b$ in frame $a$.
We also assume that the LiDAR frame coincides with the ego-vehicle frame for convenience. 
Therefore, the pose of the ego-vehicle in frame $W$ at $t_k$ is represented as $\mathbf{T}^W_{L_k}$ and the pose
transformation from $t_{k-1}$ to $t_k$ is represented as $\mathbf{T}^{L_{k-1}}_{L_k}$.
For simplification, we denote $\mathbf{T}^W_{L_k}$ as ${\mathbf{T}^W_k}$ and $\mathbf{T}^{L_{k-1}}_{L_k}$ as $\mathbf{T}^{k-1}_{k}$.
In addition, the pose of $j$-th object in ${W}$ and the ego-vehicle frame at $t_k$ are represented as $\mathbf{T}^{W}_{k,O_{j}}$ and $\mathbf{T}^{L}_{k,O_{j}}$, respectively.
Each can be converted into the other as follows:
\begin{equation}
   {\mathbf{T}^W_{k,O_{j}}} = {\mathbf{T}^W_k} \cdot \mathbf{T}^{L}_{k,O_{j}}
   \label{eq_B}
\end{equation}

An overview of the proposed system is shown in \Reffig{fig_system}. 
The system consists of four modules, Prepocessing, LiDAR Odometry, Factor Graph optimization, and Multi-object tracking, which will be detailed in the following sections.

\subsection{Preprocessing}
\label{sec_pre}
\subsubsection{Object Detection}
To simplify the object observation model, we use an open source real-time 3D LiDAR object detector CenterPoint \cite{yin2021center} to generate the 3D bounding box and pose $\mathbf{T}^{L}_{k,O_{j}}$ of an object in LiDAR frame.
\subsubsection{IMU Pre-integration}
We perform IMU pre-integration to aggregate raw IMU measurements in the local frame, following \cite{forster2016manifold}.
\subsection{Multi-object Tracking}
\label{sec_track}
We use the approach in DL-SLOT \cite{tian2022dl} to predict the position of the tracked object by fitting its trajectory in a sliding window with a cubic order polynomial.
Then, a $M$ by $N$ matching matrix $\boldsymbol{\Psi}_k$ is generated by calculating the distances between the $M$ detected objects in the current scan and the predicted positions of the $N$ tracked objects in the previous scan. 
The element $\psi_k^{i, j}$ in $\boldsymbol{\Psi}_k$ indicates the matching score between the $i$-th detected object and the $j$-th tracked object. 
It can be calculated by \Refeq{eq_score}.

\begin{equation}
   \psi_k^{i,j}=\left\{
        \begin{array}{ll}
          1 - \frac{d_{i,j}}{\alpha}  & (\ d_{i,j} < d_\mathrm{thres}) \\
        
        \specialrule{0em}{0.5ex}{0.5ex}
        0                                       & otherwise \\
        \end{array}\right.
        \label{eq_score}
\end{equation}
where $\alpha$ is a constant, $d_{i,j}$ is the distance between the detected object and the predicted position of the tracked object, and $d_\mathrm{thres}$ is the association distance threshold.
The continuous shortest path algorithm \cite{ahuja1988network} is employed to perform data association based on the matching matrix. 

After completing the data association, the average velocities of the objects within the sliding window are calculated. 
An object is determined to be dynamic when its average velocity is greater than a given threshold $v_\mathrm{thres}$.

\begin{figure}[!ht]
   \centering
   \includegraphics[width=8.0cm]{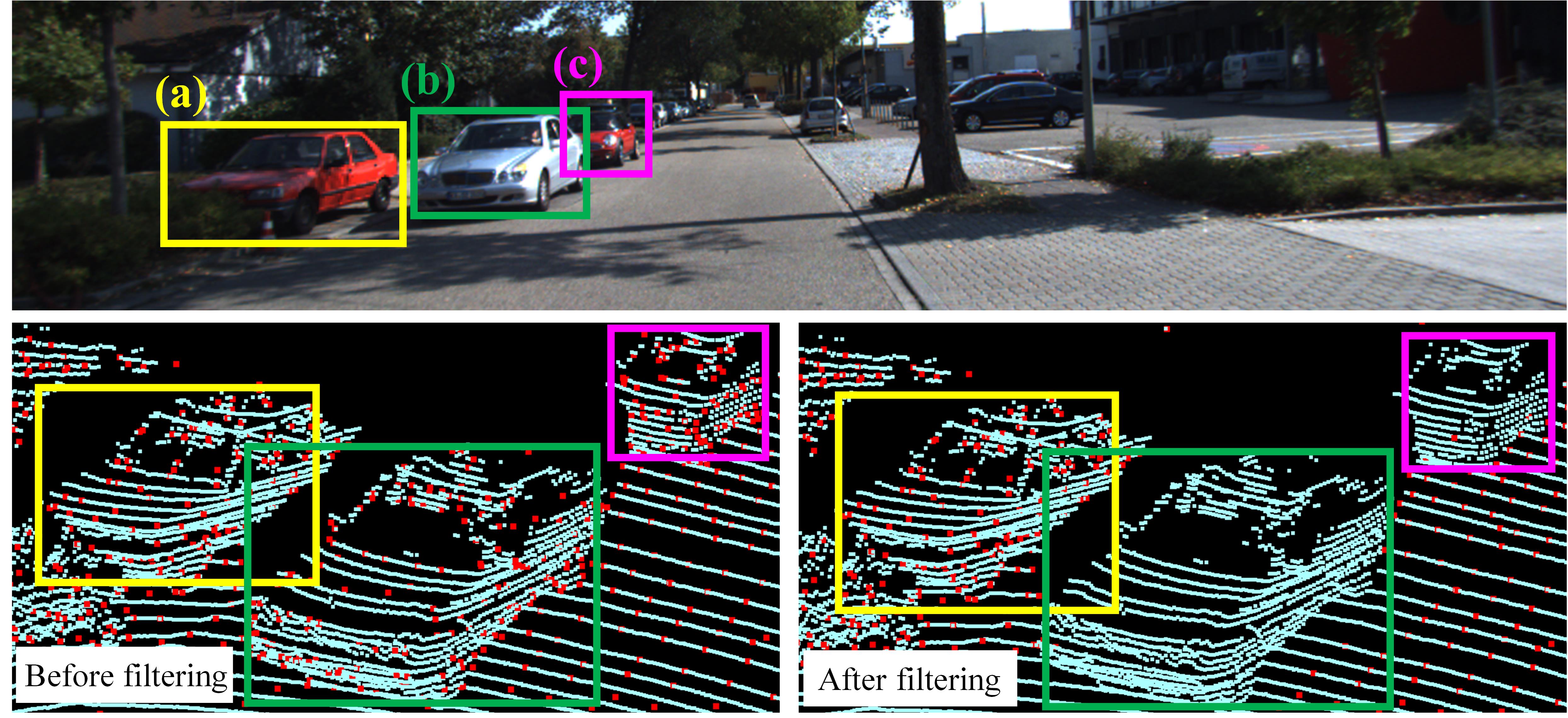}
   \caption{Example of filtering out dynamic feature points. 
   The light blue points denote the original point cloud and the red points denote the feature points. 
   The feature points on the moving cars (b) and (c) are removed exactly by LIMOT, while the feature points on the static car (a) are remained.} 
   \label{fig_filter}
\end{figure}
\subsection{LiDAR Odometry}
\label{sec_lo}
\subsubsection{De-skewing, Feature Extraction and Scan-matching}
The LiDAR Odometry module is mainly inherited from LIO-SAM.
For point cloud de-skewing, a nonlinear motion model is utilized with the estimated motion from the IMU, which is more precise than using a linear motion model \cite{ye2019tightly}.
Edge and planar feature points are then extracted based on the local roughness of the point cloud.
After removing feature points belonging to moving objects (see \Refsec{sec_filter}), LOAM-based scan-matching \cite{loam2014} is conducted between the current scan and the historical submap to obtain ego-pose $\mathbf{T}^W_k$. 
The initial transformation guess is obtained using the predicted ego-motion, $\widetilde{\mathbf{T}}^W_{k}$, from IMU pre-integration.

\subsubsection{Trajectory-based Dynamic Feature Filtering}
\label{sec_filter}
For dynamic objects at $t_{k}$, we use their fitted trajectories generated by the Multi-object tracking module to predict their positions at $t_{k+1}$.
Based on the predicted ego-pose $\widetilde{\mathbf{T}}^W_{k+1}$ at $t_{k+1}$, feature points located within the 3D bounding boxes of dynamic objects can be filtered out from the $k+1$-th point cloud before scan-matching.
An illustration is shown in \Reffig{fig_filter}.
There are three cars (labeled by (a), (b), and (c)) in this scene.
Car (a) is parked on the side of the road, and car (b) and car (c) are traveling on the road.
It can be seen that our method can filter out the feature points (shown in red) extracted from the moving objects and remain the feature points on static objects.

\subsection{Joint Factor Graph Optimization}
\label{sec_factorgraph}
\begin{figure}[ht]
   \centering
   \includegraphics[width=8.0cm]{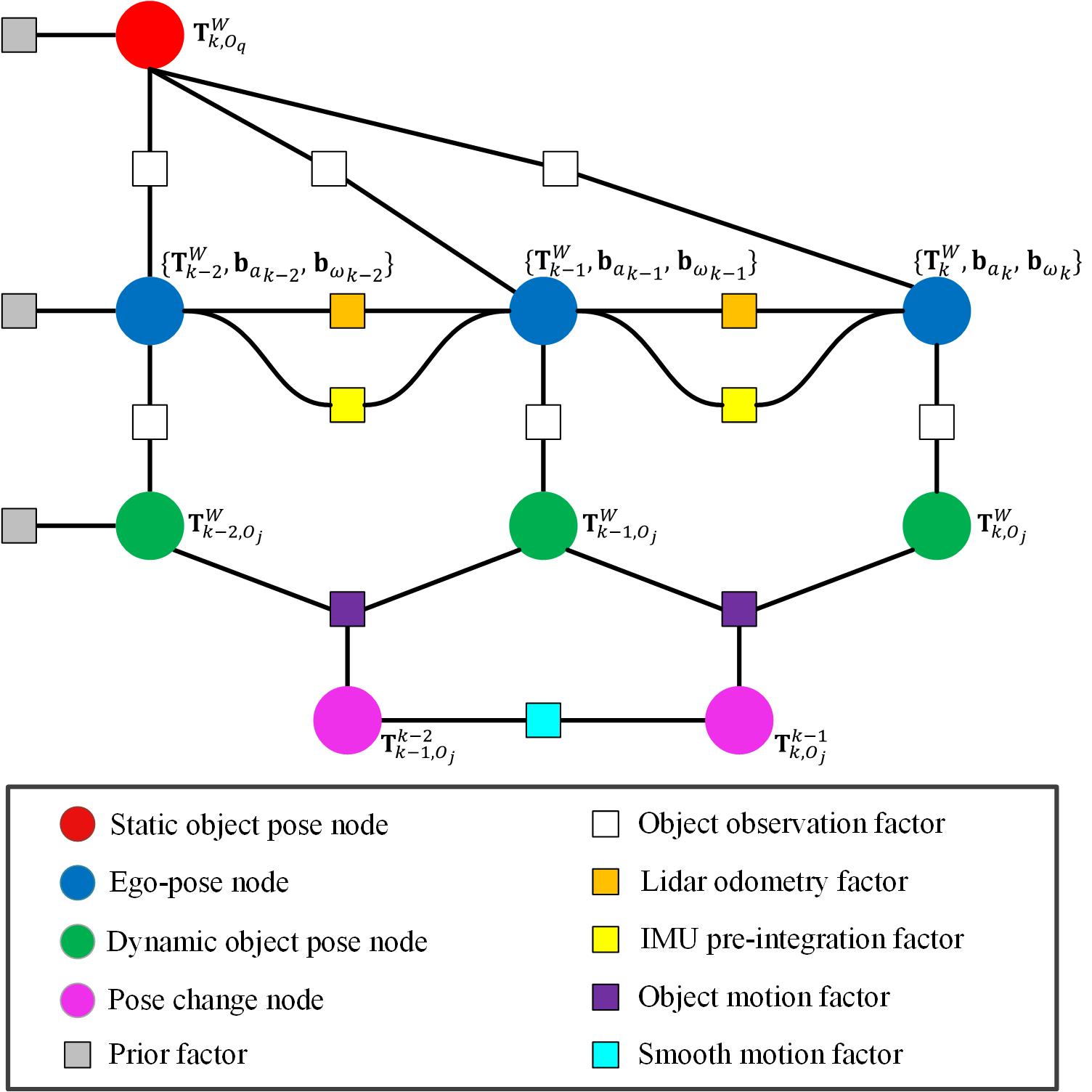}
   \caption{Factor graph framework of LIMOT for joint optimization.} 
   \label{fig_factorgraph}
\end{figure}

The joint factor graph optimization framework is shown in \Reffig{fig_factorgraph}.
It consists of the factors providing constraints for optimization and variable nodes including the states of the ego-vehicle and surrounding objects.
When an object has been tracked for 5 consecutive frames, we consider it to be initialized. 
Only the pose nodes of the initialized objects are added to the factor graph, which avoids false detections.
We add pose nodes for initialized dynamic objects at each time $t_{k}$ but maintain only one pose node in the factor graph when the object is judged as stationary to ensure the uniqueness of its global pose.
This provides reliable static observation constraints.
The residual formulation of each factor is given below.

Given the ego-poses at $t_{k-1}$ and $t_k$ estimated by LiDAR scan-matching, the residual of the LiDAR odometry factor can be defined as follows:
\begin{equation}
   \mathbf{r}_{odo}^{k}(\mathbf{T}^{W}_{k-1},{\mathbf{T}^{W}_{k}})=({\mathbf{T}^{W}_{k-1}}^{-1} \cdot {\mathbf{T}^{W}_{k}})\cdot {\mathbf{T}^{k-1}_{k}}^{-1}
   \label{eq_r_odo}
\end{equation}
Then, based on the object detection results, the residual of the object observation factor is defined as follows:
\begin{equation}
   \mathbf{r}_{obs}^{k,j}({\mathbf{T}^{W}_{k}},{\mathbf{T}^{W}_{k,O_j}}) = ( {{\mathbf{T}^{W}_{k}}^{-1}} \cdot {\mathbf{T}^{W}_{k,O_j}}) \cdot {\mathbf{T}^{L}_{k,O_j} }^{-1}
   \label{eq_r_obs}
\end{equation}
The pose change of object ${\mathbf{T}^{k-1}_{k,O_j}}$ between $t_{k-1}$ and $t_k$ can be calculated using \Refeq{eq_moti}, which also indicates the velocity of the object.

The object motion factor is a ternary factor associated with three nodes: two dynamic object pose nodes ${\mathbf{T}^{W}_{k-1,O_j}}$, ${\mathbf{T}^{W}_{k,O_j}}$ and one pose change node ${\mathbf{T}^{k-1}_{k,O_j}}$.
These are continually updated throughout the optimization process, but they must always satisfy \Refeq{eq_moti}.
Thus, the residual of the object motion factor is defined by \Refeq{eq_r_moti}.
\begin{equation}
   {\mathbf{T}^{k-1}_{k,O_j}} = {\mathbf{T}^{W}_{k-1,O_j}}^{-1} \cdot {\mathbf{T}^{W}_{k,O_j}}
   \label{eq_moti}
\end{equation}
\begin{equation}
   \mathbf{r}_{moti}^{k,j}({\mathbf{T}^{W}_{k-1,O_j}},\!{\mathbf{T}^{W}_{k,O_j}},\!{\mathbf{T}^{k-1}_{k,O_j}})\!=\!
   ({\mathbf{T}^{W}_{k-1,O_j}}\!^{-1} \!\cdot \!{\mathbf{T}^{W}_{k,O_j}}) \!\cdot \!{\mathbf{T}^{k-1}_{k,O_j}}^{-1}
    \label{eq_r_moti}
\end{equation}
We assume that a dynamic object moves at a constant velocity over a short period of time, such that the pose changes of the object at successive times should be almost identical.
Therefore, the residual of the smooth motion factor is defined as:
\begin{equation}
   \mathbf{r}_{smoo}^{k,j}({\mathbf{T}^{k-2}_{k-1,O_j}},{\mathbf{T}^{k-1}_{k,O_j}})= {\mathbf{T}^{k-2}_{k-1,O_j}}^{-1} \cdot  {\mathbf{T}^{k-1}_{k,O_j}}
        \label{eq_r_smoo}
\end{equation}
\begin{table*}[htb!]
   \caption{The RMSE of ATE$_{\mathrm{T}}$[m] and ATE$_{\mathrm{R}}$[rad] Results of Ego-pose Estimation Comparison on the KITTI Tracking Dataset.
   Bold and Underlined Text Indicate the Best and the Suboptimal Result, Respectively}
   \centering
   \setlength{\tabcolsep}{0.7mm}{ 
   \begin{tabular}{c|c c|c c|c c|c c|c c|c c}
      \toprule 
      \multicolumn{1}{c|}{\multirow{2}{*}{Seq}} & \multicolumn{2}{c|}{LIO-SAM \cite{shan2020lio}}  & \multicolumn{2}{c|}{DL-SLOT \cite{tian2022dl}}& \multicolumn{2}{c|}{LIO-SEGMOT \cite{lin2023lio-segmot}} & \multicolumn{2}{c|}{LIO-Allfilt}& \multicolumn{2}{c|}{LIO-Dynafilt}& \multicolumn{2}{c}{LIMOT} \\ 
      \multicolumn{1}{c|}{}  & ATE$_{\mathrm{T}}$[m]    & ATE$_{\mathrm{R}}$[rad]       & ATE$_{\mathrm{T}}$[m]    & ATE$_{\mathrm{R}}$[rad]     & ATE$_{\mathrm{T}}$[m]    & ATE$_{\mathrm{R}}$[rad]    & ATE$_{\mathrm{T}}$[m]    & ATE$_{\mathrm{R}}$[rad] & ATE$_{\mathrm{T}}$[m]    & ATE$_{\mathrm{R}}$[rad] & ATE$_{\mathrm{T}}$[m]    & ATE$_{\mathrm{R}}$[rad]\\ 
      \midrule
      00 & 0.856 & {\ul0.024} & 0.984 & 0.031 & 0.867 & \textbf{0.020} & 0.854 & {\ul0.024} & {\ul0.852} & {\ul0.024} & \textbf{0.846} & {\ul0.024} \\
      01 & {\ul1.683} & {\ul0.044} & 1.764 & 0.193 & 1.704 & \textbf{0.043} & 1.824 & \textbf{0.043} & 1.686 & {\ul0.044} & \textbf{1.681} & 0.045 \\
      02 & 0.274 & {\ul 0.013} & 0.362 & 0.023 & \textbf{0.253} & 0.016 & 0.280 & 0.014 & 0.274 & \textbf{0.012} & {\ul0.269} & {\ul0.013} \\
      03 & \textbf{0.245} & 0.031 & 1.920 & 0.050 & 0.269 & 0.038 & 0.247 & \textbf{0.028} & {\ul0.246} & \textbf{0.028} & \textbf{0.245} & {\ul0.030} \\
      04 & 0.660 & {\ul0.104} & 1.050 & 0.276 & 0.657 & 0.114 & 0.659 & 0.116 & {\ul0.634} & \textbf{0.099} & \textbf{0.629} & 0.107 \\
      05 & 0.347 & 0.134 & 1.209 & \textbf{0.100} & \textbf{0.313} & 0.124 & 0.346 & 0.127 & {\ul0.342} & 0.122 & 0.355 & {\ul0.112} \\
      06 & 0.266 & 0.013 & 1.925 & \textbf{0.081} & 0.202 & {\ul0.012} & {\ul0.199} & {\ul0.012} & 0.206 & {\ul0.012} & \textbf{0.190} & {\ul0.012} \\
      07 & 1.319 & {\ul0.025} & 1.759 & 0.272 & 1.376 & 0.032 & 1.331 & {\ul0.025} & {\ul1.308} & \textbf{0.024} & \textbf{1.278} & \textbf{0.024} \\
      08 & 1.274 & {\ul0.307} & 2.807 & \textbf{0.284} & \textbf{1.120} & \textbf{0.284} & 1.253 & 0.328 & 1.257 & 0.310 & {\ul1.237} & 0.311 \\
      09 & 1.201 & \textbf{0.024} & 1.597 & 1.123 & \textbf{1.135} & {\ul0.025} & 1.181 & \textbf{0.024} & 1.220 & {\ul0.025} & {\ul1.175} & 0.030 \\
      10 & {\ul0.438} & {\ul0.108} & 0.860 & 0.137 & 0.495 & \textbf{0.106} & 0.447 & 0.118 & {\ul0.438} & \textbf{0.106} & \textbf{0.425} & \textbf{0.106} \\
      11 & {\ul0.262} & 0.202 & 0.303 & 0.214 & 0.282 & 0.283 & \textbf{0.260} & 0.182 & {\ul0.262} & \textbf{0.150} & 0.266 & {\ul0.181} \\
      13 & 0.260 & 0.044 & 0.331 & 0.051 & {\ul0.258} & \textbf{0.040} & 0.263 & {\ul0.043} & 0.262 & {\ul0.043} & \textbf{0.256} & 0.045 \\
      14 & {\ul0.105} & {\ul0.013} & 0.151 & 0.197 & \textbf{0.095} & {\ul0.013} & 0.111 & 0.014 & 0.107 & {\ul0.013} & 0.108 & \textbf{0.012} \\
      15 & 0.275 & 0.066 & 0.385 & \textbf{0.012} & 0.291 & 0.076 & {\ul0.274} & {\ul0.065} & \textbf{0.273} & 0.066 & {\ul0.274} & 0.066 \\
      18 & 0.385 & \textbf{0.141} & 0.851 & 0.166 & \textbf{0.371} & 0.168 & 0.374 & {\ul0.157} & 0.376 & 0.159 & {\ul0.373} & 0.160 \\
      19 & {\ul0.916} & {\ul0.122} & 1.006 & 0.358 & 0.955 & 0.136 & {\ul0.916} & 0.131 & {\ul0.916} & 0.135 & \textbf{0.915} & \textbf{0.120} \\
      20 & 9.488 & \textbf{0.021} & 22.349 & 0.043 & 9.313 & {\ul0.023} & 1.863 & 0.027 & {\ul1.834} & 0.025 & \textbf{1.710} & 0.027 \\
      \midrule
      mean & 1.125 & 0.080 &2.312 & 0.201& 1.109 & 0.086 & 0.705 & 0.082 & {\ul0.694} & \textbf{0.078} & \textbf{0.680} & {\ul0.079}  \\
      \bottomrule
      \end{tabular}}
      \label{tab_ego_track}
\end{table*}
Finally, the optimization problem can be denoted as:
\begin{equation}
   \begin{aligned}
      \mathcal{X}^*= & \underset{\mathcal{X}}{argmin} \Big\{ \left \| \mathbf{r}_{P}(\mathcal{X}) \right \| ^{2} + \sum_{k\in \mathcal{I}} \left \| \mathbf{r}_{I}^{k}(\mathcal{X}) \right \| ^{2}_{Q_{I}} \\
         &\!+\!\sum_{k\in \mathcal{L}} \big(\left \|  \log(\mathbf{r}_{odo}^{k}(\mathcal{X})) \right \| ^{2}_{Q_{odo}} \!+\! \sum_{j\in \mathcal{O}_{k} } (\left \| \log(\mathbf{r}_{obs}^{k,j}(\mathcal{X})) \right \| ^{2}_{Q_{obs}} \\
           &\!+\! \left \|  \log(\mathbf{r}_{moti}^{k,j}(\mathcal{X})) \right \| ^{2}_{Q_{moti}} \!+\! \left \| \log(\mathbf{r}_{smoo}^{k,j}(\mathcal{X})) \right \| ^{2}_{Q_{smoo}} )\big)
      \Big\}
   \end{aligned}
   \label{eq_r_all}
\end{equation}
where $\mathcal{X}$ is the set of all variables, $\left \|\mathbf{r}_{P}(\mathcal{X}) \right \| ^{2}$ is the prior from marginalization, $\mathcal{I}$ is the set of all IMU measurements, $\mathcal{L}$ is the set of LiDAR scans in the sliding window, and $Q$ represents the covariance matrix.
$\mathbf{r}_{I}^{k}(\mathcal{X})$ is the formula of the IMU pre-integration factor residual. 
The operation $\log(·)$ represents the transformation from SE(3) to $\mathfrak{se}$(3).
Each LiDAR scan is related to the tracked object set $\mathcal{O}_{k}$.

\section{Experiments}
We conducted experiments using the public KITTI tracking dataset \cite{Geiger2012CVPR} and the self-collected dataset to evaluate the performance of the proposed LIMOT. 
All experiments were based on real-time detection results from CenterPonit, whose weights are trained on the Argoverse 2 dataset \cite{wilson2023argoverse}.
For all the experiments, we set the sliding window size to 5, $v_\mathrm{thres}$ to 1 m/s, and $\alpha$ to 100. 
Additionally, the $d_\mathrm{thres}$ for initialized object association was set to 2 m. 
Experiments were carried out on a workstation with Ubuntu 20.04, equipped with an Intel Core Xeon(R) Gold 6248R 3.00GHz processor, 32G RAM, and a NVIDIA RTX A4000 16GB graphic card.

\subsection{Evaluation Metrics}
The root-mean-square error (RMSE) of the translational and the rotational absolute trajectory errors (ATE$_{\mathrm{T}}$ [m] and ATE$_{\mathrm{R}}$ [rad]) are adopted to assess the accuracy of ego-poses \cite{zhang2018tutorial}.
The performance of multi-object tracking is evaluated in two ways following \cite{bescos2021dynaslam}. 
The pose accuracy of the objects is also evaluated using the RMSE of ATE$_{\mathrm{T}}$ [m] and ATE$_{\mathrm{R}}$ [rad].
The multi-object tracking precision (MOTP) metric \cite{bernardin2008evaluating} is used to evaluate tracking precision. 
MOTP results are only given on the KITTI tracking dataset since it provides the ground truth tracking results for evaluation.

\begin{figure*}[ht]
   \centering
   \includegraphics[width=17.0cm]{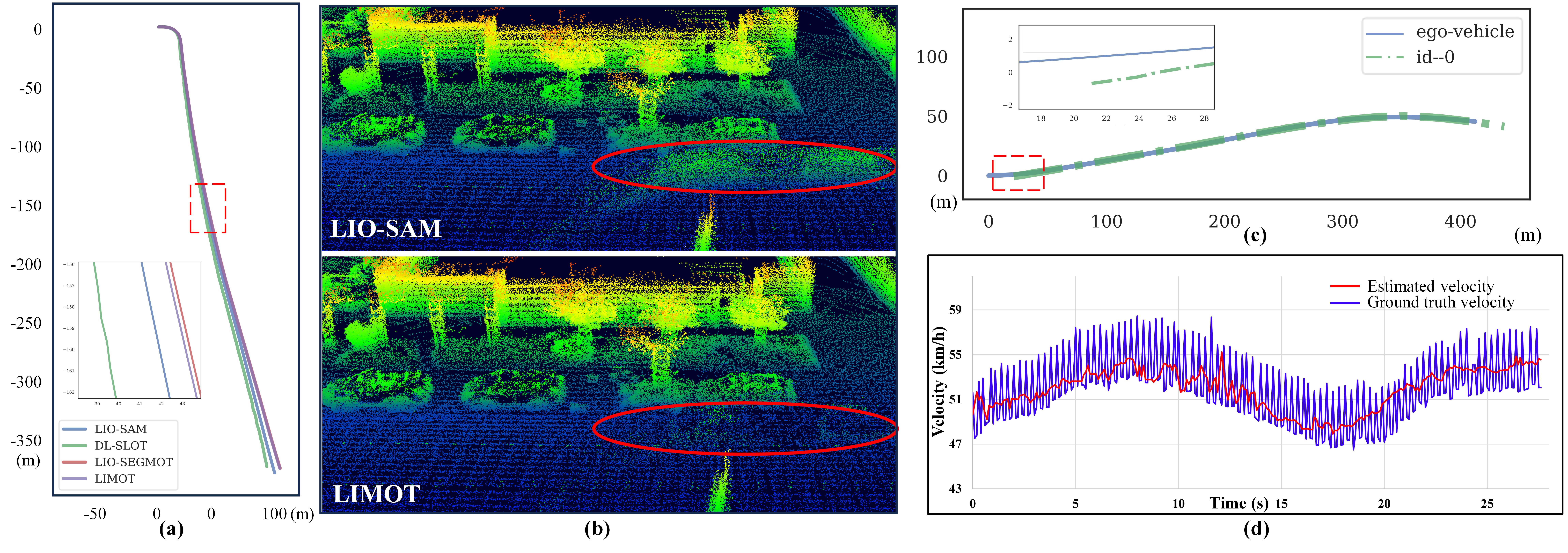}
   \caption{Qualitative results of LIMOT on the KITTI tracking dataset. 
   (a) Comparison of ego-trajectories in sequence 04 of the KITTI tracking dataset.
   (b) Comparison of point cloud maps generated by LIO-SAM and LIMOT. 
   (c) Trajectories of the main tracked object (id 0) and ego-vehicle in sequence 10 of the KITTI tracking dataset.
   (d) Comparison between the ground truth and estimated instantaneous velocity of the tracked object in (c).}
   \label{fig_qualitative}
\end{figure*}

\subsection{Baselines}
The baseline multi-object tracking and SLAM methods are DL-SLOT \cite{tian2022dl} and LIO-SEGMOT \cite{lin2023lio-segmot}.
LIO-SAM \cite{shan2020lio} is also compared since it is one of the state-of-the-art (SOTA) LiDAR-inertial SLAM methods, on which we build LIMOT. 
When only the proposed dynamic feature point removal is performed without joint optimization, the method is referred to as LIO-Dynafilt.
In addition, we also provide experimental results of removing all feature points on movable objects without joint optimization, LIO-Allfilt.

We selected two popular 3D multi-object tracking methods, AB3DMOT \cite{weng2020ab3dmot} and PC3T \cite{wu20213d}, as well as DL-SLOT and LIO-SEGMOT as the baseline multi-object tracking methods. 
AB3DMOT first proposed to directly evaluate multi-object tracking in 3D space, which is suitable for LiDAR-based approaches. 
PC3T is a SOTA multi-object tracking method that requires ground truth ego-poses as input.

\subsection{KITTI Tracking Dataset}
\subsubsection{Ego-pose Evaluation}
The KITTI tracking dataset was collected in urban areas and along highways. 
We chose all the sequences in the KITTI tracking dataset except for sequences without ego-motion.
The comparative results are shown in \Reftab{tab_ego_track}.
LIO-Allfilt shows a large translational error in sequence 01 because parked cars are representative in this sequence.
The filtering of all their feature points impedes scan-matching.
The pose accuracy of LIO-Dynafilt is higher than those of LIO-SAM and LIO-Allfilt, demonstrating the effectiveness of our proposed dynamic feature filtering approach.
Compared to DL-SLOT, all other methods achieve better performance, demonstrating that coupling IMU assists in localization.
LIMOT achieves the best pose estimation performance in many of the evaluated sequences, confirming that jointly optimizing the poses of the ego-vehicle and objects is beneficial.
LIO-SEGMOT does not improve the pose accuracy much compared to LIO-SAM because it does not consider the effect of dynamic features on the scan-matching.
It is important to note that sequence 20 is a highway scene containing a large number of moving objects, so LIMOT, LIO-Dynafilt, and LIO-Dynafilt all have a large improvement compared to LIO-SAM and LIO-SEGMOT.
Moreover, since the highway is a typical scene for LiDAR SLAM degradation, DL-SLOT shows the largest translational error.

The comparison of ego-trajectories of sequence 04 is shown in \Reffig{fig_qualitative} (a).
The point cloud maps of sequence 01 generated by LIO-SAM and LIMOT are shown in \Reffig{fig_qualitative} (b).
In the red ellipse of the former, there is a very obvious ghosting caused by the moving objects, which, however, is eliminated from the point cloud generated by LIMOT.
\begin{table}[tb]
   \caption{MOTP [\%] Result Comparison of Different Multi-object Tracking Algorithms on the KITTI Tracking Dataset}
   \centering
   \setlength{\tabcolsep}{0.7mm}{ 
   \begin{tabular}{c|cc}
   \toprule 
   \multicolumn{1}{c|}{\multirow{2}{*}{Method}} & \multicolumn{2}{c}{MOTP {[}\%{]}} \\
   \multicolumn{1}{c|}{} & \multicolumn{1}{c}{(IOU$_\mathrm{thres}$   = 0.25)} & \multicolumn{1}{c}{(IOU$_\mathrm{thres}$ = 0.5)} \\
   \midrule
   AB3DMOT & 61.37 & 64.17 \\[0.4ex]
   PC3T & 61.26 & 64.07 \\[0.4ex]
   DL-SLOT & 60.13 & 63.60 \\[0.4ex]
   LIO-SEGMOT & 61.45 & 64.22 \\[0.4ex]
   LIMOT & \textbf{62.25} & \textbf{64.26}\\[0.4ex]
   \bottomrule
   \end{tabular}
   }
   \label{tab_ab3dmot_compare}
\end{table}
\begin{table*}[ht]
   \caption{Results of Object Pose Estimation Comparison on the KITTI Tracking Dataset. Bold and Underlined Text Indicate the Best and the Suboptimal Result, Respectively}
   \centering
   \setlength{\tabcolsep}{0.7mm}{ 
   \begin{tabular}{c|c c c|c c c|c c c}
      \toprule 
      \multicolumn{1}{c|}{\multirow{2}{*}{Seq / Obj.id}}  & \multicolumn{3}{c|}{DL-SLOT \cite{tian2022dl}}& \multicolumn{3}{c|}{LIO-SEGMOT \cite{lin2023lio-segmot}} & \multicolumn{3}{c}{LIMOT} \\ 
      \multicolumn{1}{c|}{}    &TP [\%]     & ATE$_{\mathrm{T}}$[m]    & ATE$_{\mathrm{R}}$[rad]  &TP [\%]   & ATE$_{\mathrm{T}}$[m]    & ATE$_{\mathrm{R}}$[rad]  &TP [\%] & ATE$_{\mathrm{T}}$[m]    & ATE$_{\mathrm{R}}$[rad] \\ 
      \midrule
      04 / 2             &{\ul27.07}      &2.799 & 1.760 &\textbf{27.71} &{\ul0.711}	&{\ul0.241}	&	{\ul27.07}&	\textbf{0.697}	&\textbf{0.233}      \\ 
      05 / 31           &\textbf{40.40}      & 0.924& 0.223   &{\ul 40.07}&	{\ul0.304}&	{\ul0.175}&	\textbf{40.40}&	\textbf{0.302}&	\textbf{0.163}      \\ 
      08 / 8              &{\ul23.33}& 0.661  & 0.346 & \textbf{24.36}&	{\ul0.430}&	\textbf{0.122}&		23.08&	\textbf{0.384}&	{\ul0.130}     \\ 
      08 / 13             &{\ul48.85}&0.827 &0.116 & \textbf{50.76} &	{\ul0.656}&	\textbf{0.104}	&	{\ul48.85}&	\textbf{0.641}	&{\ul0.105}      \\ 
      10 / 0             &\textbf{93.88}&0.943 & {\ul0.168}  &{\ul88.44}&	{\ul0.439}	&\textbf{0.156	}&	\textbf{93.88} &\textbf{0.368	} & \textbf{0.156}      \\ 
      11 / 0            &\textbf{51.21}&0.445 &\textbf{0.129} & {\ul50.40}&	{\ul0.292}&	{\ul0.186}		&\textbf{51.21}&	\textbf{0.277} &	0.189      \\ 
      18 / 2            &29.17&0.566 &0.878  & \textbf{31.44} & \textbf{0.246} &	\textbf{0.580}	&	{\ul30.68}&	{\ul0.261}	&{\ul0.602}     \\ 
      18 / 3            &{\ul46.67}&0.493 &\textbf{0.456} & 45.61&	{\ul0.386}&	0.560	&	\textbf{47.02} &	\textbf{0.372}	&{\ul0.459}      \\ 
      20 / 12           &37.79&10.342 & 0.106  & {\ul45.10}&	{\ul1.701}&	{\ul0.105}	&	\textbf{46.66}&	\textbf{0.826}&	\textbf{0.102}     \\ 
      20 / 122            &\textbf{30.20}&1.005 & \textbf{0.117} & {\ul29.41}&	\textbf{0.258}&	{\ul0.127}	&	\textbf{30.20}&	{\ul0.292}	&0.131      \\ 
      \midrule
      \multicolumn{1}{c|}{mean}   & 42.86&	1.901&	0.430&43.33&	0.542&	0.236	&	\textbf{43.90}&	\textbf{0.442}&	\textbf{0.227}\\     
      \bottomrule
      \end{tabular}
      }
      \label{tab_obj_track}
\end{table*}

\subsubsection{Tracking Precision Evaluation}
The comparison results are shown in \Reftab{tab_ab3dmot_compare}.
For each tracked object, its Intersection over Union (IoU) with ground truth label should exceed a threshold IoU$_\mathrm{thres}$ to be considered as a successful match.
It can be found that our method shows the best results in terms of MOTP, indicating that our method achieves better tracking precision for all tracked objects.

\subsubsection{Object Pose Evaluation}
We further evaluated the pose accuracy of the individual objects. 
We selected 10 objects with the longest tracking frame length in \Reftab{tab_ego_track} for evaluation. 
The experimental results are shown in \Reftab{tab_obj_track}.
Here TP [\%] stands for the ratio of the number of tracked frames to the number of the frames of the object's ground truth trajectory.
The IoU threshold for evaluating TP is taken as 0.25.
Since we adopt the tracking method in DL-SLOT, the TP results of these two methods are the same on most sequences.
However, the pose accuracy of objects is improved significantly in LIMOT, largely due to the reduction of the ego-pose rotation error by coupling IMU measurements.
Moreover, LIMOT and LIO-SEGMOT track the objects for almost the same number of frames, but LIMOT achieves better pose accuracy of the objects.

The trajectories of the tracked object (id 0) and ego-vehicle in sequence 10 are demonstrated in \Reffig{fig_qualitative} (c).
We further compare the estimated instantaneous velocity of the object (id 0) in sequence 10 with the ground truth.
As shown in \Reffig{fig_qualitative} (d), the ground truth instantaneous velocities (blue line) exhibit significant fluctuation due to the imprecise object annotations within the dataset.
In contrast, the estimated instantaneous velocities by LIMOT (red line) are smoother and in the middle of the fluctuation range of the ground truth.
This indicates that it is appropriate to add the constant velocity constraint to the factor graph optimization. 

\begin{figure}[tb]
   \centering
   \includegraphics[width=8.0cm]{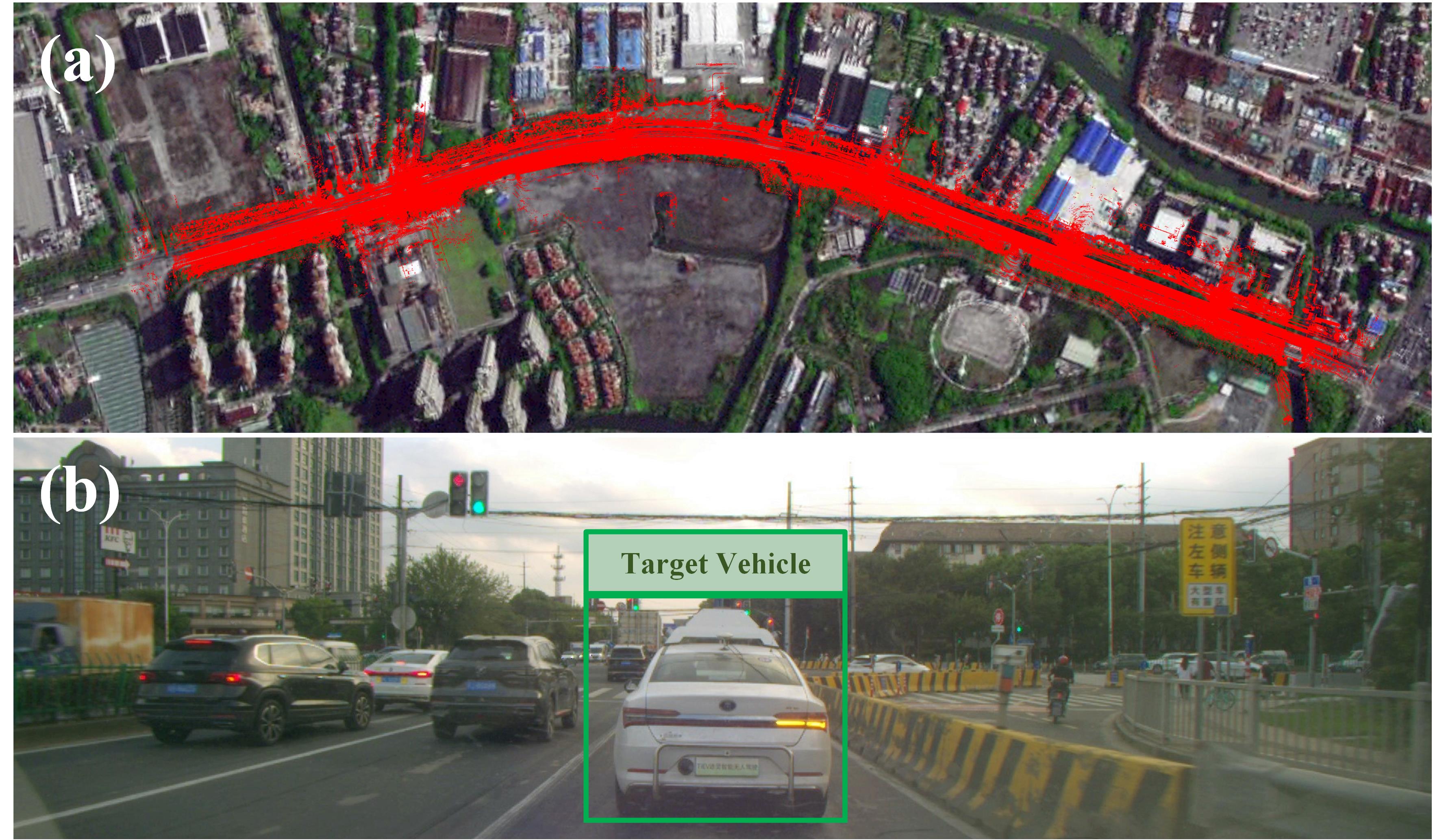}
   \caption{The overview of the self-collected dataset. 
   (a) LIMOT mapping result aligning with the satellite map. 
   (b) A representative front view image of the self-collected dataset.}
   \label{fig_jiading}
\end{figure}
\subsection{Self-collected Dataset}
\subsubsection{Data Collection}
We collected this dataset using the TIEV platform \cite{zhao2018tiev} on the North Jiasong Road, Jiading District, Shanghai.
\Reffig{fig_jiading} shows an overview of this dataset and it can be seen that this scene is rich in dynamic objects.
TIEV is equipped with a Velodyne HDL-64E LiDAR and an RTK-Inertial Navigation System (INS), which provides IMU data at 100 HZ as well as high-precision ground truth.
In addition, we used a target vehicle, playing the role of the tracked object, equipped with the same INS, so we can obtain its ground truth trajectory.

\subsubsection{Pose Evaluation}
Both the ego-vehicle and the target vehicle pose evaluation results are shown in \Reftab{tab_jiading}.
Compared to LIO-SAM, the ego-pose result of LIMOT presents in an average improvement of 10.14\% and 6.12\% in terms of ATE$_{\mathrm{T}}$ and ATE$_{\mathrm{R}}$, respectively.
As for the target object pose result, although LIO-SEGMOT tracks the object for a longer period of time, it shows a significant translational error.
LIMOT obtains the most accurate target object position.
\begin{table}[tb]
   \caption{Results of Pose Estimation Comparison on the Self-collected Dataset.
   Bold and Underlined Text Indicate the Best and the Suboptimal Result, Respectively}
   \centering
   \setlength{\tabcolsep}{0.7mm}{ 
   \begin{tabular}{c|c c|  |c c c}
      \toprule 
      \multicolumn{1}{c|}{\multirow{2}{*}{Method}}  & \multicolumn{2}{c|  |}{Ego-pose}& \multicolumn{3}{c}{Target object pose} \\ [0.4ex]
      \multicolumn{1}{c|}{} &  ATE$_{\mathrm{T}}$[m]    & ATE$_{\mathrm{R}}$[rad]  &TP [\%]   & ATE$_{\mathrm{T}}$[m]    & ATE$_{\mathrm{R}}$[rad] \\
      \midrule
      LIO-SAM & 3.344 & 0.049 & -- & -- & -- \\[0.4ex]
      DL-SLOT & 5.188 & 0.118 & 68.61 & 0.524 & {\ul0.092} \\[0.4ex]
      LIO-SEGMOT & {\ul3.243} & \textbf{0.046} & \textbf{85.99} & 2.593 & \textbf{0.079} \\[0.4ex]
      LIO-Allfilt & 3.358 & \textbf{0.046} & -- &--  & -- \\[0.4ex]
      LIO-Dynafilt & 3.314 & {\ul0.047} & -- & -- & -- \\[0.4ex]
      LIMOT & \textbf{3.005} & \textbf{0.046} & {\ul71.73} & \textbf{0.212} & {\ul0.092} \\
      \bottomrule
      \end{tabular}
      }
      \label{tab_jiading}
\end{table}
\subsection{Running Time Analysis}
We calculated the average time-consumption of the main functional modules of LIMOT on the KITTI tracking dataset.
The results are shown in \Reftab{tab_time_consume}.
Compared with LIO-SAM, the time consumption of LiDAR odometry and factor graph optimization increase by 0.4 ms and 7.4 ms, respectively, which correspond to the runtime of the dynamic feature filtering and joint optimization of the states of objects. 
LIO-SEGMOT performs significantly slower than LIMOT because of its computationally intensive factor graph structure.
\begin{table}[tb]
   \caption{Average Time-consuming of the Main Functional Modules for Processing One Scan}
   \centering
   \setlength{\tabcolsep}{0.7mm}{ 
   \begin{tabular}{c|c|c|c}
      \toprule 
   Module & \multicolumn{3}{c}{Average Runtime (ms)} \\
   \midrule 
   LiDAR Odometry & \multicolumn{3}{c}{78.4 + 0.4} \\[0.4ex]
   Multi-object   Tracking & \multicolumn{3}{c}{0.9} \\[0.4ex]
   Factor Graph   Optimization & \multicolumn{3}{c}{1.5 + 7.4} \\
   \midrule 
    \midrule 
    & LIO-SAM & LIO-SEGMOT & LIMOT \\
    \midrule
   FPS & 12.6 & 7.5 & 11.3 \\
   \bottomrule
   \end{tabular}
   }
   \label{tab_time_consume}
   \end{table}
\section{CONCLUSIONS}
We present LIMOT, a tightly-coupled system for LiDAR-inertial SLAM and multi-object tracking capable of jointly optimizing the poses of the ego-vehicle and objects in a sliding window.
Furthermore, this method can filter out feature points on moving objects based on the approximated object trajectories, while the remaining feature points on static objects are used to provide constraints for scan-matching, which enhances the robustness of the system and improves its performance in dynamic environments.
Experimental results show that LIMOT improves the pose accuracy of ego-vehicle and objects, as well as the tracking accuracy, which demonstrates that LiDAR-inertial SLAM and multi-object tracking can exhibit mutual benefits on each other.
Future work could involve introducing a dynamics model for moving objects in the environment to obtain more accurate object states. 

\bibliographystyle{IEEEtran}
\bibliography{reference}
\end{document}